\UseRawInputEncoding
\documentclass[letterpaper, 10 pt, conference]{ieeeconf}  % Comment this line out if you need a4paper

\usepackage{graphicx}
\usepackage{epsfig} % for postscript graphics files
\usepackage{booktabs}
\usepackage{multirow}
\usepackage{times} % assumes new font selection scheme installed
\usepackage{amsmath,amssymb,amsfonts} % assumes amsmath package installed
\usepackage{cleveref}
\usepackage{threeparttable}
\usepackage{array}

\IEEEoverridecommandlockouts                              % This command is only needed if 
\title{\LARGE \bf
Leveraging Pre-trained Large Language Models with Refined Prompting for Online Task and Motion Planning
}

\author{Huihui Guo, Huilong Pi, Yunchuan Qin, Zhuo Tang, Kenli Li% <-this % stops a space
\thanks{Huihui Guo is with the Hunan University, Changsha, 410082 China (e-mail: ghh1991@hnu.edu.cn). }
\thanks{Huilong Pi is with the Hunan University, Changsha, 410082 China (e-mail: phl880217@163.com). }
\thanks{Yunchuan Qin is with the Hunan University, Changsha, 410082 China (e-mail: qinyunchuan@hnu.edu.cn). }
\thanks{Zhuo Tang is with the Hunan University, Changsha, 410082 China (e-mail: ztang@hnu.edu.cn). }
\thanks{Kenli Li is with the Hunan University, Changsha, 410082 China (e-mail: lkl@hnu.edu.cn). }
}

\begin{document}

\maketitle
\thispagestyle{empty}
\pagestyle{empty}

%%%%%%%%%%%%%%%%%%%%%%%%%%%%%%%%%%%%%%%%%%%%%%%%%%%%%%%%%%%%%%%%%%%%%%%%%%%%%%%%
\begin{abstract}
With the rapid advancement of artificial intelligence, there is an increasing demand for intelligent robots capable of assisting humans in daily tasks and performing complex operations.
Such robots not only require task planning capabilities but must also execute tasks with stability and robustness.
In this paper, we present a closed-loop task planning and acting system, LLM-PAS, which is assisted by a pre-trained Large Language Model (LLM).
While LLM-PAS plans long-horizon tasks in a manner similar to traditional task and motion planners, it also emphasizes the execution phase of the task.
By transferring part of the constraint-checking process from the planning phase to the execution phase,
LLM-PAS enables exploration of the constraint space and delivers more accurate feedback on environmental anomalies during execution.
The reasoning capabilities of the LLM allow it to handle anomalies that cannot be addressed by the robust executor.
To further enhance the system's ability to assist the planner during replanning, we propose the First Look Prompting (FLP) method, which induces LLM to generate effective PDDL goals.
Through comparative prompting experiments and systematic experiments,
we demonstrate the effectiveness and robustness of LLM-PAS in handling anomalous conditions during task execution.

\end{abstract}

%%%%%%%%%%%%%%%%%%%%%%%%%%%%%%%%%%%%%%%%%%%%%%%%%%%%%%%%%%%%%%%%%%%%%%%%%%%%%%%%
\section{INTRODUCTION}

The ability of robots to autonomously execute long-horizon tasks in unstructured environments is a key milestone in advancing robotic autonomy.
Task and Motion Planning (TAMP) is a promising approach for planning long-horizon tasks, as it effectively integrates low-level geometric information to guide robots through complex task execution  \cite{b1, b2}.
While TAMP can generate action sequences with sufficient geometric information,
significant challenges remain in ensuring the successful execution of these actions in real-world environments, particularly in scenarios that deviate from the closed-world assumption (CWA) \cite{b2.1}.

Traditional control-based task planning methods incorporate real-time environmental feedback during execution, closing the loop between planning and execution.
These online planners replan or take corrective actions when a failure occurs due to an anomaly or dynamic change \cite{b3, b4, b5}.
Typically, these methods address anomalies by manually predefined recovery strategies or heuristic rules.
However, setting recovery strategies requires domain experts to design problem-specific rules, which are difficult to generalize.
Furthermore, certain tasks, such as searching for a missing object, demand active exploration by the agent and cannot be readily solved by heuristic planners.
Besides, many online planners focus on the planning process and assume idealized anomaly detection, relying on accurate sensor data from simulated scenarios or complete visual localization coverage \cite{b4, b6}.
In practice, some anomalies necessitate that the robot fully explore its constraint space to provide meaningful feedback.
For instance, consider a scenario where the robot is tasked with grasping an object on a table, but the object is not detected in the initially targeted area.
This type of anomaly is common in dynamic environments. In such cases, the robot may need to explore other regions of the table to locate the object,
and if unsuccessful, then it can provide more accurate anomaly feedback that is "the target object is not on the table".
This feedback can then be utilized by the robot to adjust its strategy accordingly.
% Thus, to ensure the successful execution of long-horizon tasks in unstructured environments, the TAMP planner must generate a comprehensive plan that guides the robot's actions.
% Equally important is the ability to monitor the execution process to acquire precise environmental data.

\begin{figure}[!t]
        \centerline{\includegraphics[width=1.0\columnwidth]{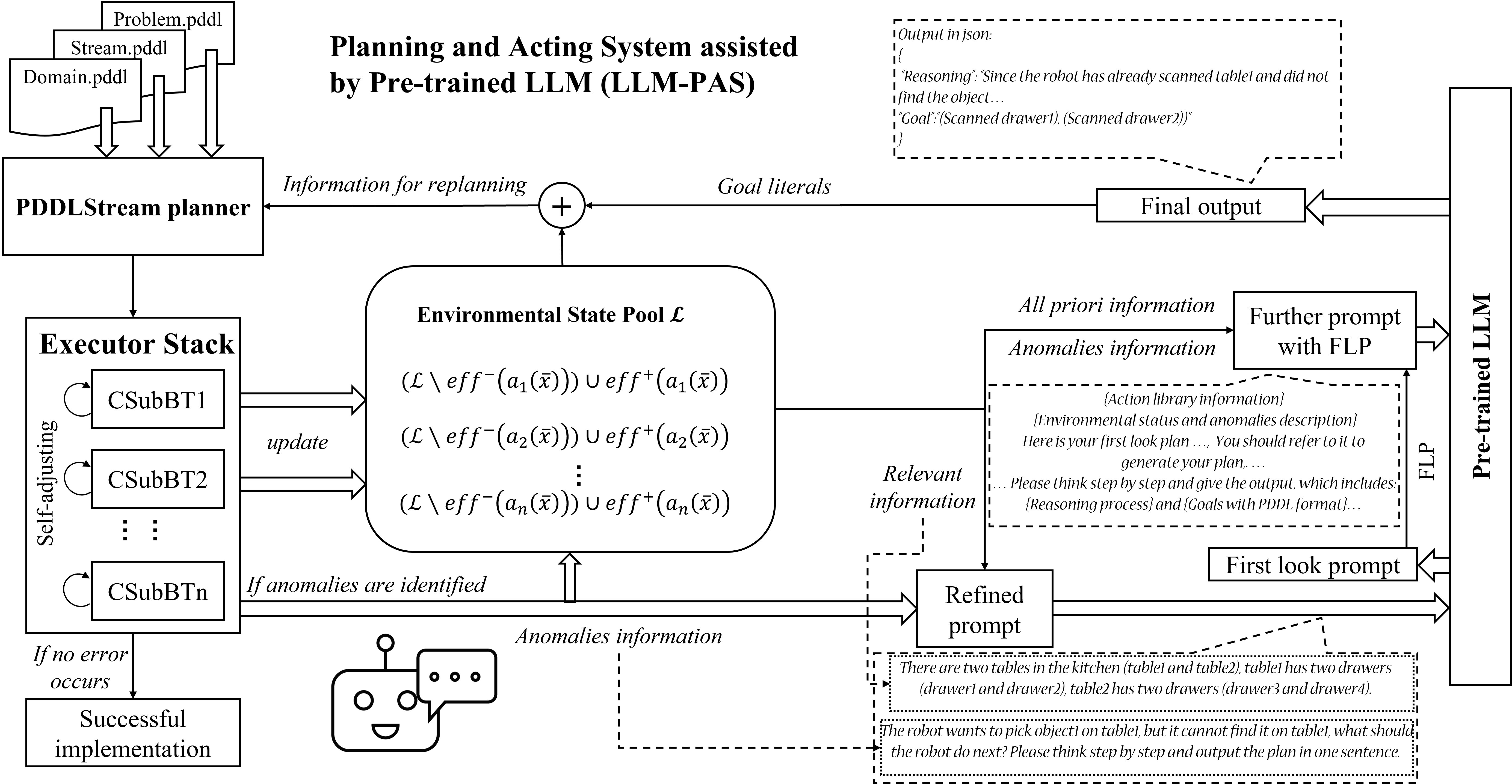}}
        \caption{Overview of LLM-PAS. PDDLStream planner generates TAMP solutions,
                which are subsequently converted into CSubBTs for execution. When CSubBTs encounter unsolvable anomalies, they provide feedback on the abnormal information.
                This anomaly information, along with other relevant context, is sent to the LLM module using FLP.
                The LLM module then processes this information and outputs new PDDL task goals to facilitate the replanning process.}
        \label{fig_framework}
\end{figure}

When anomalous information is detected, the planner's effectiveness can be significantly enhanced if it possesses human-like logical reasoning capabilities.
This allows it to address potential anomalies across various tasks, thereby expanding its applicability and robustness in handling unexpected situations.
LLMs have shown impressive logical reasoning capabilities in task planning without additional parameter fine-tuning \cite{b7, b8}.
LLMs have also been applied in TAMP tasks, generating continuous geometric parameters when provided with suitable in-context prompts \cite{b9}.
However, for more complex planning problems, such as long-horizon TAMP tasks, LLMs are still insufficient to fully replace domain experts and specialized planners \cite{b10}.
% Additionally, LLMs require intricate and precise prompt engineering to produce detailed and reliable plans.
Thus, directly substituting LLMs for classic task planners in guiding robots through complex TAMP tasks remains unreliable and difficult to generalize.

% Thus, directly substituting LLMs for classic task planners in guiding robots through complex TAMP tasks remains unreliable and difficult to generalize.
However, from a different perspective, LLMs can effectively translate natural language instructions into formal task descriptions, such as PDDL formats \cite{b10, b11}.
Leveraging LLMs to assist traditional TAMP planners, for instance, by providing PDDL goals, is a feasible approach.
% LLMs exhibit human-like reasoning, enabling them to analyze and propose solutions based on environmental conditions when encountering anomalies.
The general logical reasoning capabilities of LLMs can compensate for the shortcomings of traditional TAMP planners, particularly in addressing exceptions more flexibly.

In our previous work, we transformed the actions from the TAMP solution into conditional subtrees of Behavior Trees \cite{b12} (CSubBTs) for execution,
which provide robust execution capability and constrained space exploration capability.
Building on this foundation, we propose a closed-loop system for planning and acting, assisted by LLM (LLM-PAS).
The system generates action sequences through a traditional TAMP planner and automatically converts these actions into CSubBTs for robust execution.
Perturbations during execution, such as sensor errors, can be resolved through internal adjustments within the CSubBTs.
However, if an anomaly arises that cannot be handled by the CSubBTs, the system will pass the unsatisfiable constraint information to the LLM module after fully exploring the constraint space.

To ensure the LLM module focuses on addressing exceptions without being distracted by irrelevant context, we introduce a method called First Look Prompting (FLP).
In FLP, only the anomaly-related information is provided as the initial prompt for the LLM module.
The LLM's output is then used as the FLP, and a subsequent prompt is created by adding relevant contextual information.
The final output of the LLM module is formatted as PDDL goals for the planner to conduct online replanning in response to anomalies.

In summary, our contributions are as follows:

\begin{itemize}
        \item We propose a planning and acting system with robust performance and generalizable anomaly reasoning capabilities for TAMP applications.
        \item We introduce the FLP method and demonstrate, through comparisons with other LLM-based reactive planners in typical anomaly scenarios, that it generates more effective and detailed PDDL goals, aiding planners in replanning.
        \item We validate the effectiveness of the LLM-PAS system through both simulation and real-world experiments, showing its ability to handle common anomalies and successfully complete tasks.
\end{itemize}

\section{Related Work}

\subsubsection{Task and Motion Planning}
TAMP integrates low-level motion planning with high-level task planning to generate feasible plans for long-horizon tasks.
The planning problem is formalized using domain-specific symbolic languages, such as PDDL, where the plan contains detailed geometric information that guides the robot in task execution.
Sampling-based TAMP approaches reduce the dimensionality of the sampling space by leveraging action constraints to search for feasible solutions \cite{b13}.
The sampled geometric information is transformed into certified domain symbolic representations, which are then used as preconditions for the TAMP planner's actions \cite{b14}.

However, these methods primarily focus on planning and often overlook challenges that arise during execution.
Online TAMP methods address this by considering replanning in response to anomalies or unpredictable state changes \cite{b4, b6}.
The robustness of plan execution is typically managed by searching the belief space or employing appropriate control strategies.
Nonetheless, these approaches often rely on predefined domain knowledge or rules, making them difficult to generalize to new scenarios.

Our work aims to fully integrate planning and acting, being able to explore the constraint space during execution and then provide feedback when anomalies arise.
We leverage the general logical reasoning capabilities of LLMs to handle anomalies,
eliminating the need for complex domain rule modeling or replanning strategies, while being able to guarantee that the task resumes normal execution.

\subsubsection{Reactive Planning with LLMs}
The advent of pre-trained LLMs has significantly impacted the field of artificial intelligence.
Among these, transformer-based LLMs \cite{b15, b19, b15.2, b21} have become valuable tools for robotic planning in the AI planning community \cite{b10}.
Few-shot or one-shot prompting \cite{b16, b17} allows LLMs to generate action sequences similar to classic planners without requiring additional training \cite{b7, b18}.
Some methods also account for unpredictable situations by using LLMs to handle exceptions.
For example, InnerMonologue \cite{b15.1} leverages the reasoning ability of LLMs and environmental feedback, such as object detection, scene description, and success/failure detection, to form a closed-loop planning system.
Similarly, PROGPROMPT \cite{b20} continuously checks code-like assertions to obtain feedback from the environment and generate corrective actions,
leveraging the fact that LLMs are trained on programming languages, thus representing actions as Pythonic functions.
However, both of these methods impose significant formal requirements on prompt engineering when querying and responding to the LLMs. 
In comparison to our method, they struggle to generate correct outputs in the absence of suitable prompting examples for abnormal problems. 
As a result, they lack the ability to generalize in complex tasks and environments.
Additionally, these approaches typically rely on LLMs for task-level logic planning,
without fully integrating low-level geometric constraints, making them difficult to apply directly to TAMP tasks.

CoPAL \cite{b15.3} proposes a multi-layer feedback planning framework that includes low-level motion information.
While multiple levels of error feedback enhance the completeness of the planning system, 
the complex cross-relationships of LLMs at different levels can lead to uncontrollable final outputs, particularly in complex problems.
Additionally, LLMs are not yet capable of fully replacing traditional planners \cite{b10}, especially in TAMP applications. 
In this paper, we propose using LLMs to assist traditional TAMP planners in replanning, 
enabling the planner to leverage general reasoning capabilities for more effective exception handling.

% However, these approaches typically rely on LLMs for task-level logic planning,
% without fully integrating low-level geometric constraints, making them difficult to apply directly to TAMP tasks.

% Text2Reaction \cite{b21} considers additional environmental changes that may cause the current plan to fail.
% This method enables LLMs to analyze situations in a step-by-step manner using specialized re-planning prompts, facilitating the creation of new plans to handle complex reactive planning challenges.

% For example, CAPE \cite{b19} detects whether artificially predefined preconditions for an action are met, triggering exception handling when necessary.
% The LLM-based reasoning module is then informed of the error type and generates corrective actions to recover the original plan.

% 第二章整体思路应该首先介绍TAMP相关工作，指出以前的TAMP工作要么假设CWA，要么假设传感器能够理想反馈信息，
% 接着提出规划的执行也同样重要，需要深入的跟进，并且指出一些环境反馈信息也是需要通过探索约束空间才能准确得到的。
% 然后介绍LLM对于任务规划领域的作用,指出LLM并不能很好的胜任规划器的工作,尤其是TAMP问题,进而提出通过LLM来协助规划器完成重规划.

\section{Background}

% This work aims to address TAMP challenges for robotic manipulation tasks in dynamic unstructured environments.
% Building upon the PDDLStream TAMP planner \cite{b14}, we extend its application to dynamic environments without assuming a CWA.
% To execute the sequence of actions generated by the TAMP planner, we employ Behavior Trees (BTs).
% As demonstrated in our previous work, the generated BTs can self-adjust during execution to accommodate tolerable deviations or disturbances.
% In cases where an unsolvable anomaly arises, the system fully explores the constraint space before reporting the final error information.
% This approach ensures that the environmental feedback obtained is both accurate and contextually relevant.

\subsection{Problem Formulation with PDDLStream}

PDDLStream \cite{b14} is a TAMP framework built on top of PDDL, which allows procedural sampling of continuous values in the form of {\itshape streams}.
Similar to PDDL, PDDLStream uses {\itshape predicate} logic to formalize planning problems.
A {\itshape predicate p} is a Boolean function, and its evaluation for a set of arguments is referred to as a {\itshape literal}.
If the Boolean function returns true, the {\itshape literal} becomes a {\itshape fact}; otherwise, it is considered a {\itshape negated fact}.

In PDDLStream, there are two types of {\itshape literals}.
{\itshape Fluent literals} can change their truth value when arguments fail to satisfy certain constraints.
These literals appear both in the preconditions and effects of {\itshape actions}, such as predicates like {\itshape AtBConf} and {\itshape AtPose}.
On the other hand, {\itshape static literals} represent constant facts and are generated by low-level motion constraints.
For example, the predicate {\itshape (Kin ?a ?o ?p ?g ?t)} denotes that the robot's arm (?a) can plan a trajectory (?t) with grasp (?g) for object (?o) at pose (?p) when the robot is in state (?q).
Additionally, types such as {\itshape (Arm ?a)}, {\itshape (Grasp ?g)}, {\itshape (Object ?o)}, {\itshape (Pose ?p)},
{\itshape (State ?q)}, and {\itshape (ATraj ?t)} are treated as {\itshape unary predicates}, which are also considered {\itshape static literals} and correspond to the parameters of actions.

% 考虑加入一个表，表示用了哪些Static literals和Fluent literals

\begin{figure}[!t]
        \centerline{\includegraphics[width=\columnwidth]{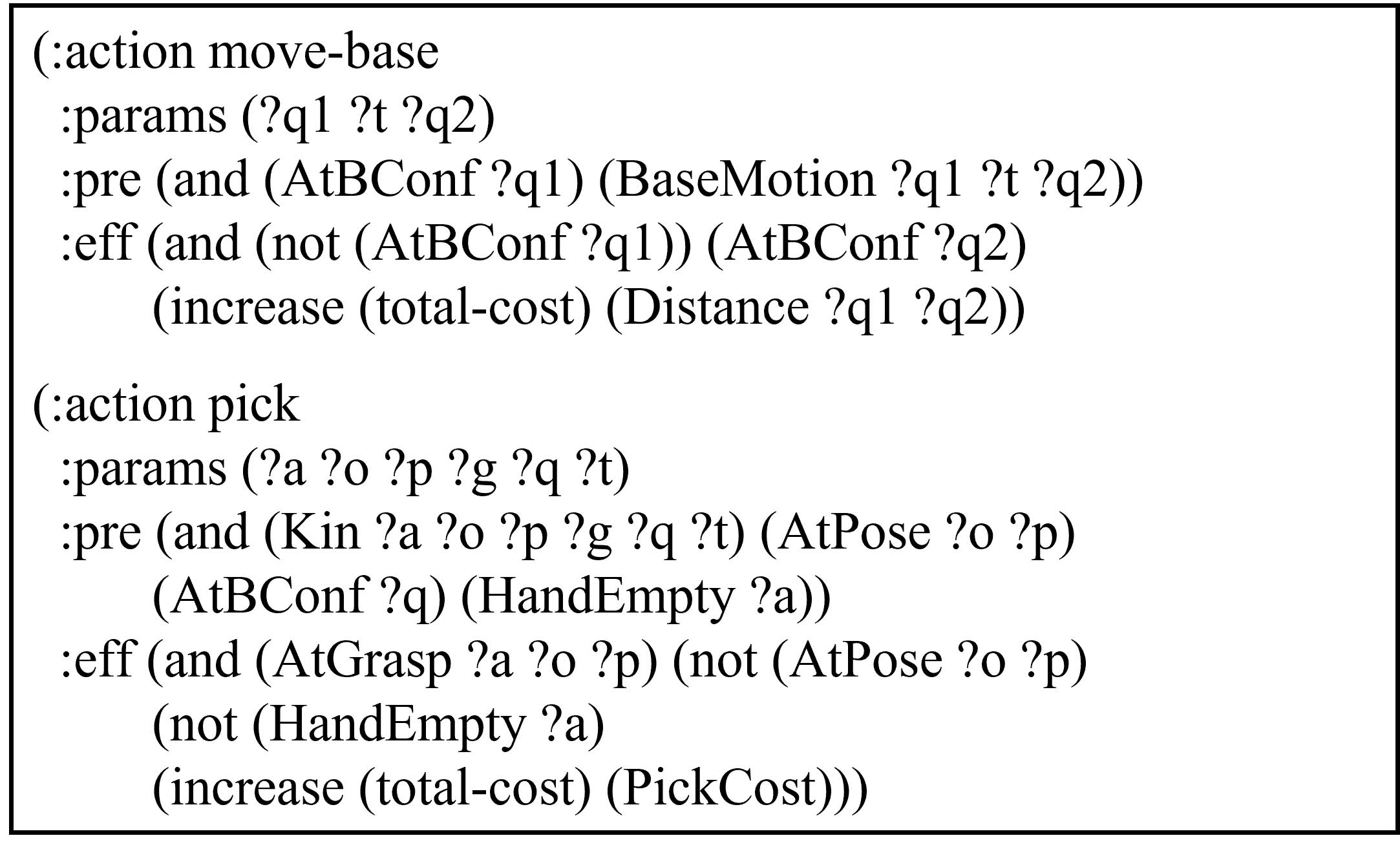}}
        \caption{Two example actions of the domain.pddl file.
                The preconditions incorporate the {\itshape static literals} ({\itshape BaseMotion} and {\itshape Kin}).}
        \label{fig_action_eg}
\end{figure}

\begin{figure}[!t]
        \centerline{\includegraphics[width=\columnwidth]{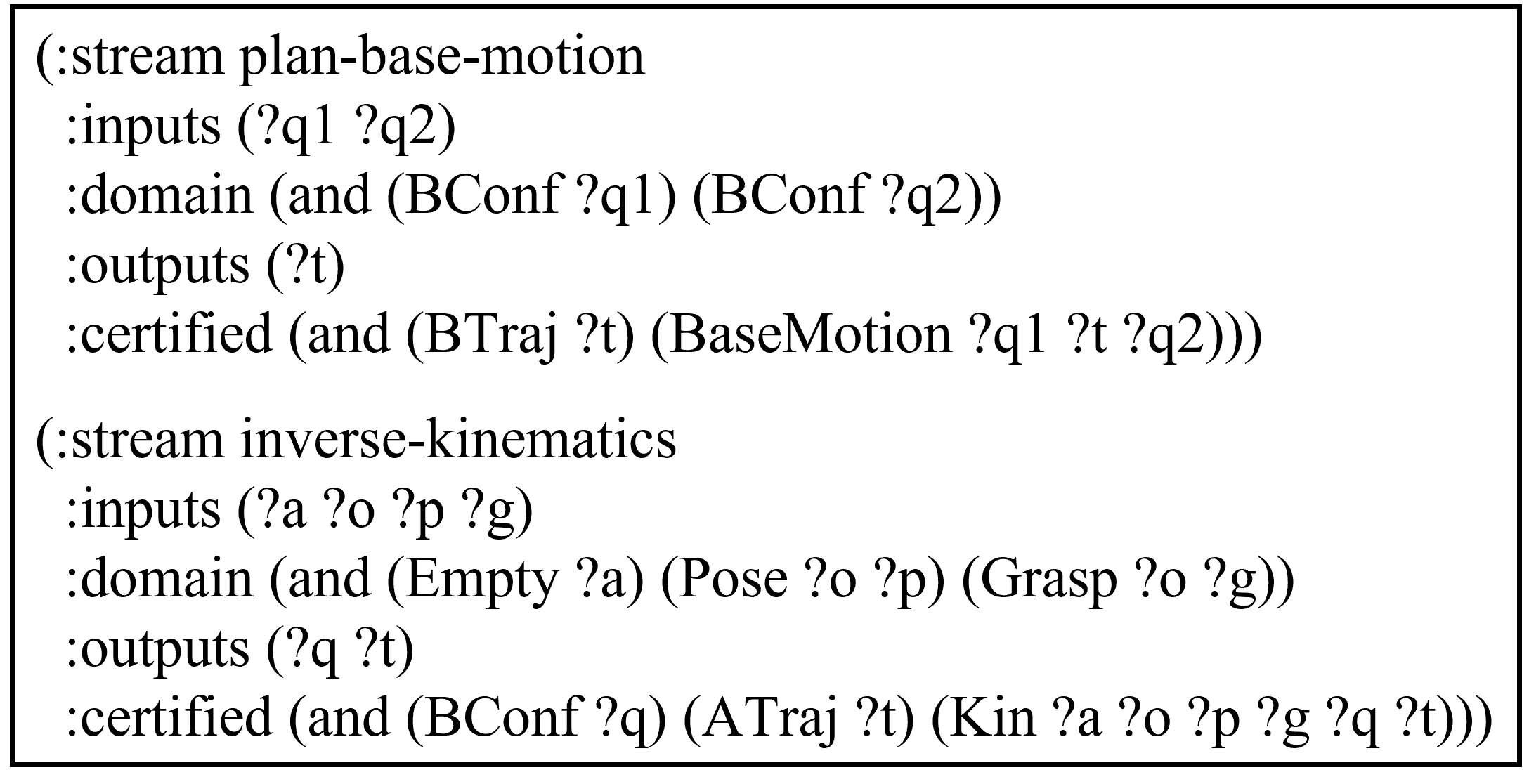}}
        \caption{Two example streams of the stream.pddl file.
                The streams generate the {\itshape static literals} ({\itshape BaseMotion} and {\itshape Kin}).}
        \label{fig_stream_eg}
\end{figure}

The representation of {\itshape actions} in PDDLStream is similar to those in PDDL,
with the key difference being that {\itshape static literals} are included in the preconditions of the {\itshape actions}, as shown in \Cref{fig_action_eg}.
These {\itshape static literals} encapsulate low-level geometric information, which is generated by the {\itshape streams}.
Unlike conventional PDDL planning, PDDLStream introduces an additional stream.pddl file to define stream-related information.
Each {\itshape stream} typically corresponds to a {\itshape conditional sampler} $\psi = \left\langle I, O, \mathcal{C}, f \right\rangle $,
which samples geometric values from the intersection of the constraint space, and searches for the control parameters required by the action.
In \Cref{fig_stream_eg}, the parameters in (\textbf{:inputs}) and (\textbf{:outputs}) represent the inputs $I$ and outputs $O$ of the {\itshape conditional sampler},
while the (\textbf{:domain}) and (\textbf{:certified}) formulas define the structure of the inputs and outputs.
The $\mathcal{C}$ in $\psi$ corresponds to the constraints of the {\itshape stream}, and $f$ represents the sampling function.

When the PDDLStream planner searches for a plan based on the domain information,
the {\itshape streams} retrieve the relevant {\itshape literals} provided by the sampling function,
following the search direction (whether backward or forward) until a feasible solution is identified.

% In this work, we assume that some of these constraints are satisfied by default during the planning phase,
% while transfering each conditional sampler into the execution phase to instantly confirm that these constraints are satisfied during the execution process,
% and if not, to facilitate direct adjustments in the execution phase.
% To illustrate, given the inherent uncertainty associated with the execution of a given plan,
% the {\itshape BaseMotion} and {\itshape Kin} constraints are more reliable
% to be evaluated in the immediate period preceding implementation rather than in the planning phase without closed-world assumptions.
% Furthermore, by re-positioning this process in the execution phase, the planner can reduce the overall planning time.b

\subsection{Goal Configurations for Planning}

In PDDLStream, an action instance $a(\bar{x})$ is applicable in a state $\mathcal{I}$ if $(pre^+(a(\bar{x})) \subseteq \mathcal{I}) \wedge (pre^-(a(\bar{x})) \cap \mathcal{I} = \emptyset )$,
where the $+$ and $-$ superscripts correspond to the positive and negative {\itshape literals}, respectively.
After applying the action instance, the new state $\mathcal{I}$ is updated as $(\mathcal{I}\setminus eff^-(a(\bar{x}))) \cup eff^+(a(\bar{x}))$.
A PDDL problem consists of a set of goal {\itshape literals}. Given a set of {\itshape actions} and an initial state $\mathcal{I}$,
if the planner finds a sequence of applicable {\itshape actions} that achieves the goal {\itshape literals}, then the sequence forms a feasible plan $\pi = [a_1(\bar{x}_1), \dots, a_n(\bar{x}_n)]$.

Before specifying goals in PDDLStream,
it is essential to focus on a type of {\itshape predicate} that typically does not contain geometric information or can be implicitly described by others that do ({\itshape derived predicates}).
For example, effects such as {\itshape (Cooked ?o)} and {\itshape (Cleaned ?o)} are predicates without geometric information,
while {\itshape derived predicates} like {\itshape (On ?o ?r)} and {\itshape (Holding ?a ?o)} are implicit effects of actions such as {\itshape place} and {\itshape pick}.
In this paper, we refer to these as {\itshape logical state predicates}.
The set of goal {\itshape literals} is composed by selecting appropriate {\itshape logical state predicates}.

\subsection{Conditional Subtrees of BTs}

In our previous work \cite{b14.1}, the actions in a plan $\pi$ could be directly transformed into {\itshape conditional subtrees} of Behavior Trees (CSubBTs) for execution.
CSubBT operates as a {\itshape factored transition system} \cite{b13} composed of multiple atomic actions.
For example, the {\itshape pick} action can be factored into three atomic actions: {\itshape Pre-approach}, {\itshape Approach} and {\itshape Grasp}.
The {\itshape Pre-approach} atomic action enables the manipulator to choose either a top-grasp or a forward-grasp.
The {\itshape Approach} atomic action adjusts the manipulator's end pose in Cartesian space based on the detected target pose. 
The {\itshape Grasp} atomic action completes the pick process.
Additionally, the {\itshape move} action can serve as an atomic action within the {\itshape pick} action, forming a {\itshape move-and-pick} action.
These atomic actions collectively contribute to the final {\itshape pick} action.

By integrating {\itshape conditional samplers} \cite{b13}, CSubBT adjusts the sampled parameters associated with atomic actions and manages their repeated execution.
This capability allows CSubBT to internally backtrack and modify the execution of atomic actions in response to abnormalities, 
without altering the structure of the overall plan.
Consequently, CSubBT enhances robustness and enables exploration of the associated constraint space. 
This leads to a higher success rate in executing planned actions and provides accurate feedback when encountering unresolvable anomalies.

% Additionally, several consecutive actions could be combined into a single CSubBT, such as combining {\itshape move} and {\itshape pick} into a {\itshape move-and-pick} action.
% This combination allows CSubBT to internally backtrack and adjust the execution of actions to handle abnormalities without altering the structure of the overall plan.
% A CSubBT serves as the executor of an action or a combined action, and includes the action's {\itshape conditional samplers}.

% CSubBT operates as a {\itshape factored transition system} \cite{b13} composed of multiple atomic actions.
% By integrating {\itshape conditional samplers}, CSubBT adjusts the sampled parameters associated with atomic actions and manages their repeated execution.
% This provides both robustness and the ability to explore the associated constraint space.
% CSubBT enhances the success rate of executing planned actions and offers accurate feedback when encountering unresolvable anomalies.

\section{Method}

% A comprehensive planning and acting system requires not only robust execution mechanisms but also the capability to accurately perceive and process environmental information,
% thereby forming a complete closed-loop system.
% This ensures that anomalies, which cannot be resolved by the executors, can be effectively addressed through replanning, enabling the system to adapt and respond to unforeseen challenges.

\subsection{System overview}

We propose LLM-PAS, a closed-loop planning and acting system that leverages the logical reasoning capabilities of LLMs and incorporates self-adjustment and replanning features.
The LLM-PAS framework is illustrated in \Cref{fig_framework}.
The system utilizes domain files (Domain.pddl, Stream.pddl, and Problem.pddl) to define the initial action library and related prior information.
Using these inputs, the PDDLStream planner generates a feasible plan $\pi =[a_1, a_2, \dots, a_n]$, consisting of a sequence of symbolic actions derived from the action library.

LLM-PAS automatically converts these symbolic actions or new combined actions into CSubBTs for execution.
The plan is implemented as $\pi =[A_1, A_2, \dots, A_m]$, where $m \leqslant n$ and each action $A$ is represented by a corresponding CSubBT.
These CSubBTs are added to the executor stack and executed sequentially.

Given that certain constraints cannot be fully determined until execution, LLM-PAS shifts part of the constraint-checking responsibilities from the PDDLStream planner to the CSubBTs.
For instance, constraints such as {\itshape (BaseMotion ?q1 ?t ?q2)} and {\itshape (Kin ?a ?o ?p ?g ?q ?t)} pertain to the combined {\itshape move-and-pick} action.
During the planning phase, the variable $?q2$ is the goal state of the {\itshape move} action, and is sampled from the intersection of the constraint manifold based on geometric information.
This variable $?q2$ is also used as $?q$ in the {\itshape Kin} constraint.
However, due to sensor inaccuracies, the {\itshape move} action might not precisely reach the pose $?q2$,
making it more practical to verify the constraints of the {\itshape pick} action only after completing the {\itshape move} action.
Thus, in the planning stage, all constraints involving variables like $?q2$ are assumed to be satisfied.

In the execution phase, CSubBTs verify whether constraints (e.g., {\itshape BaseMotion} and {\itshape Kin}) for its current atomic action are met based on real-time sensory data.
If constraints are not satisfied, the corresponding conditional samplers attempt alternative values.
Once the CSubBTs obtain feasible solutions and control parameters (such as the trajectory parameter $?t$) the actions can be executed successfully.

Each CSubBT is equipped with self-adjustment capabilities, enabling it to explore the constraint space of the corresponding action, thus increasing the likelihood of successful execution.
With appropriately tuned sampling parameters, a CSubBT will only conclude that a constraint is unsatisfiable after thoroughly exploring all possible values within the sampling space.
When encountering anomalies that cannot be resolved autonomously, the CSubBT provides accurate, detailed feedback on the failure, enhancing the system's ability to respond and support replanning.
The anomaly feedback, along with relevant environmental information, is used as input to the pre-trained LLM module.
% The LLM module assists the PDDLStream planner in generating a revised plan to address the encountered anomalies.

\subsection{First Look Prompting and Plan Correction}

LLM-PAS is designed to enhance the execution success rate of the initial plan $\pi$.
When an abnormal state $\mathcal{I}_m^*$ is detected by CSubBT, the current execution is suspended.
LLM-PAS then generates a repair plan $\pi^* =[A_1^*, A_2^*,\dots, A_k^*]$ based on the anomaly information and resumes execution from the modified plan until the system returns to the normal state $\mathcal{I}_m$.

To assist in generating the repair plan, we introduce the First Look Prompting (FLP) method.
Although LLMs are not ideally suited for planning long-horizon tasks, particularly those like TAMP that involve geometric constraints,
they demonstrate proficiency in problem analysis, logical reasoning, and the translation of textual input, such as the conversion of planning prompts into PDDL format.
In this approach, the anomaly information is used as the initial prompt for the LLM module.
We adopt a Chain-of-Thought (CoT) and few-shot approaches \cite{b22,b23,b16,b17} to generate detailed strategy outputs.
Initially, the LLM module is tasked solely with providing a logically reasoned solution, free of extraneous contextual information.
This solution, referred to as the FLP, is then combined with environmental prior information to form a refined prompt for the LLM module.

This refined prompt consists of: (1) {\itshape action library information}, (2) {\itshape environmental status and anomalies description},
(3) {\itshape first look plan}, and (4) {\itshape output format}.
The {\itshape action library information} includes a description of the symbolic action functions from the domain file,
along with the {\itshape logical state predicates} associated with successful action completion.
The anomaly information is structured around the unsatisfied constraints from CSubBT, integrated with relevant environmental data as the {\itshape environmental status and anomalies description}.
The {\itshape  first look plan} serves as the initial solution generated by the LLM.
Rather than directly using the LLM to search for feasible plans,
its reasoning capabilities are leveraged to assist the PDDLStream planner by providing a set of goal {\itshape literals} in cases where anomalies arise that cannot be resolved by CSubBT.
Consequently, the core output of the LLM module, informed by the refined prompt, is a set of {\itshape logical state predicates} derived through in-context learning.

\section{Experiment Evaluation}

We conducted a series of experiments to evaluate the performance of LLM-PAS.
First, we tested whether the LLM module can effectively generate goals in PDDL format when anomalies occur.
The effectiveness of these goals was assessed by domain experts.
We compare our prompting method with previous LLM-based reactive planners in classical planning domain scenarios, such as ALFRED \cite{b24} and blocks world, to evaluate its capability.
We also performed ablation experiments on the FLP prompting method to evaluate its reliability in guiding the LLM.
Additionally, we designed typical experimental scenarios, both in simulation and the real world, to systematically test whether LLM-PAS can successfully complete tasks under abnormal conditions.

\subsection{Goal Construction and Prompting Experiments}

% We assume that CSubBT effectively addresses tolerable positional deviations arising from sensor errors and other issues.
% For anomalies that cannot be resolved by CSubBT,
LLM module must be capable of reasoning based on the identified problems and generating valid PDDL goals to assist the TAMP planner in the replanning process.
To evaluate the LLM module's performance, we used the household robot domain, a well-established benchmark in classical planning.
We designed three typical types of anomalies that CSubBT could not resolve: (1) {\itshape Object Loss Anomaly}: The robot is unable to locate the task object.
(2) {\itshape Action Blocking Anomaly}: The action to be performed is obstructed or conflicts with other objects.
For instance, if a piece of bread needs to be heated in a microwave oven, but the microwave is already occupied with other items.
(3) {\itshape State Change Anomaly}: There is an unexpected change in the state of the object being manipulated.
For example, an apple intended to be placed in the fridge gets dirty during the process.

For the {\itshape Object Loss Anomaly} scenario, we designed 10 different situations where the LLM module needs to reason based on the existing state to locate the task object among two tables,
four drawers, and potentially under other coverings, such as an upside-down basket on the table.
We sampled 10 scenes from ALFRED and constructed 5 different tasks for each of the other two anomaly types.
ALFRED is a widely recognized benchmark for learning mappings from natural language commands and egocentric vision to action sequences for household tasks.
The specific actions and corresponding goal {\itshape literals} are detailed in \Cref{tab:actions}.
The task for the LLM module is to accurately and effectively generate PDDL goals based on the provided anomaly information.
% The 2,685 specific tasks in ALFRED, which combine various task parameters,
% fall into seven categories: {\itshape Pick\&Place}, {\itshape Stack\&Place}, {\itshape Pick Two\&Place}, {\itshape Clean\&Place}, {\itshape Heat\&Place}, {\itshape Cool\&Place}, {\itshape Examine in Light}.

% The experimental scenarios we designed can exemplify these task categories.

% To minimize the impact of unrelated actions in the action library on the LLM output,
% we manually selected actions specifically related to exception handling for the {\itshape action library information} used in the three scenarios.

\begin{table}
        \centering
        \caption{Related actions and PDDL goals for the three anomaly scenarios}
        \label{tab:actions}
        \resizebox{\linewidth}{!}{
                \begin{tabular}{c|c|c}
                        \toprule
                        Scenarioes              & Actions                                             & PDDL Goals                                                     \\
                        \midrule
                        Object Loss Anomaly     & Pick, Place, Scan\_table, Scan\_drawer, Scan\_cover & (Holding ?o), (On ?o ?r), (Scanned, ?r)                        \\
                        Action Blocking Anomaly & Pick, Place                                         & (Holding ?o), (On ?o ?r)                                       \\
                        State Change Anomaly    & Pick, Place, Cook/Clean/Heat/                       & (Holding ?o), (On ?o ?r), (Cooked ?o)/(Cleaned ?o)/(Heated ?o) \\
                        \bottomrule
                \end{tabular}
        }
\end{table}

InnerMono \cite{b15.1} and ProgPrompt \cite{b20} generate response plans in specific formats based on environmental anomaly feedback and LLM reasoning.
Both methods employ few-shot prompting to ensure that LLM provides the correct output.
In our experiments, we did not provide few-shot examples that directly related to anomalies for any of the prompting methods.
We also conduct three sets of comparative ablation experiments to evaluate the effectiveness of FLP in the replanning procedure:
(1) 1-step prompting without FLP, (2) 2-step prompting referring to FLP, and (3) 2-step prompting using FLP directly.
To assess the performance of these methods in completing all tasks, we employ two metrics: Success Rate (SR) and Average Success Path Length (ASPL).
The SR is calculated based on the effectiveness of the generated goals.
Considering that more specific and comprehensive task goals reduce the planner's workload and minimize the need to invoke the LLM module multiple times.
For instance, in the {\itshape Object Loss Anomaly} scenario,
searching multiple potential locations sequentially in one replanning process is more efficient than invoking the LLM module multiple times for individual areas.
Consequently, we further evaluate with the Average Success Path Length (ASPL) which is calculated as :
\begin{center}
        $ASPL = \frac{1}{n}\times \sum_{i = 1}^{n}\frac{\bar{L_i} }{L_{gt\_i}}\times SR_i $,
\end{center}
where $n$ is 10, 5, 5 in the three anomaly scenarioes respectively, $\bar{L_i}$ denotes the average path length over several valid prompts in $i$-th task,
and $L_{gt\_i}$ denotes the path length from ground truth.

\newcolumntype{C}[1]{>{\centering\arraybackslash}p{#1}}
% \newcolumntype{Y}{>{\centering\arraybackslash}X}
\renewcommand\arraystretch{1.5}

\begin{table}[htbp]
        \caption{Comparison results of Different Prompting methods}
        \centering
        \label{table_llm_prompt}
        \resizebox{\linewidth}{!}{
                \begin{threeparttable}
                        \begin{tabular}{c|c|c|c|c|c|c|c}% 其中，tabular是表格内容的环境；c表示centering，即文本格式居中；c的个数代表列的个数
                                \toprule %[2pt]设置线宽      
                                \multirow{3}*{Model}         & \multirow{3}*{Method}               & \multicolumn{2}{C{2cm}|}{Object Loss Anomaly} & \multicolumn{2}{C{2cm}|}{Action Blocking Anomaly} & \multicolumn{2}{C{2cm}}{State Change Anomaly}                                                 \\ %换行
                                \cline{3-8}
                                % \midrule %[2pt]  
                                ~                            & ~                                   & SR                                            & ASPL                                              & SR                                            & ASPL           & SR           & ASPL          \\
                                \midrule %[2pt]  
                                \multirow{5}*{GPT-3.5-turbo} & ProgPrompt                          & 0.61                                          & \textbf{0.878  }                                  & 0.04                                          & \#             & 0.4          & \textbf{0.75} \\
                                ~                            & InnerMono                           & 0.08                                          & \#                                                & 0.0                                           & \#             & 0.6          & 0.5 \\
                                ~                            & Ours 1-step prompting without FLP        & 0.5                                           & 0.239                                             & 0.0                                           & \#             & \textbf{1.0} & 0.5           \\
                                ~                            & Ours 2-step prompting refering to FLP    & 0.58                                          & 0.251                                             & 0.4                                           & 0.352          & \textbf{1.0} & 0.55          \\
                                ~                            & Ours 2-step prompting using FLP directly & \textbf{0.82  }                               & 0.482                                             & \textbf{0.56}                                 & \textbf{0.488} & \textbf{1.0} & 0.56 \\
                                \midrule %[2pt]  
                                \multirow{5}*{GPT-4.0}       & ProgPrompt                          & 0.76                                          & \textbf{1.0}                                      & 0.62                                          & \textbf{1.0}   & 0.58         & 0.793  \\
                                ~                            & InnerMono                           & 0.68                                          & 0.871                                             & 0.28                                          & \textbf{1.0}   & 0.64         & 0.56  \\
                                ~                            & Ours 1-step prompting without FLP        & 0.7                                           & 0.58                                              & 0.44                                          & 0.44           & \textbf{1.0} & 0.5           \\
                                ~                            & Ours 2-step prompting refering to FLP    & 0.94                                          & 0.94                                              & 0.84                                          & 0.84           & \textbf{1.0} & \textbf{1.0}  \\
                                ~                            & Ours 2-step prompting using FLP directly & \textbf{1.0}                                  & \textbf{1.0}                                      & \textbf{0.96}                                 & 0.96           & \textbf{1.0} & \textbf{1.0}  \\
                                % \midrule
                                \bottomrule %[2pt]     
                        \end{tabular}
                        \begin{tablenotes}
                                \item[] The bold entities indicate the best results, \# means that ASPL is meaningless at very low SR.
                        \end{tablenotes}
                \end{threeparttable}

        }
\end{table}

As shown in \Cref{table_llm_prompt}, 
our prompting method generally outperforms InnerMono and ProgPrompt in the absense of few-shot examples directly related to anomalies across all three scenarios,
demonstrating the superior generalization of LLM-PAS.
In some specific cases, the success rates of InnerMono and ProgPrompt are closer to those of the 1-step prompting method, 
which is significantly inferior to the 2-step prompting that using FLP directly.
Additionally, the lower success rates and higher ASPL observed in these methods suggest they are more limited by the output format rather than their ability to solve the underlying problem. 

In ablation experiments, the 1-step prompting method without FLP struggles to generate effective PDDL goals in the first two scenarios. 
In contrast, the 2-step prompting methods, especially when FLP is used directly, consistently produce better PDDL goals. 
In the third scenario, all three prompting methods reliably generate task goals, 
as the LLM module only needs to set PDDL goals for unmet states, allowing the planner to complete the remaining work. 
Additionally, the 2-step methods, particularly with FLP, provide more comprehensive and detailed task goals, 
reducing the planner's workload and the frequency of replanning.

However, in scenarios where the LLM module fails to generate effective PDDL goals, 
such as in the {\itshape Object Loss Anomaly} scenario, the LLM tends to propose repeated searches in a specific area, 
potentially leading the robot into a deadlock. 
In the {\itshape Action Blocking Anomaly} scenario, issues like syntax errors in constructed PDDL goals are common, 
even with the use of few-shot and chain-of-thought prompting methods. 
These problems are more frequent with the GPT-3.5-turbo model, 
and the quality of FLP in the {\itshape Action Blocking Anomaly} scenario is often suboptimal. 
However, using the FLP method helps alleviate these issues to some extent, depending on the quality of the FLP.

We evaluated the performance of the FLP method in the blocks world and compared it with two other LLM-based reactive planners.
The dataset used for the experiments is sourced from \cite{b15.2} and \cite{b15.3}. 
The blocks world problem involves generating a sequence of actions to achieve a goal configuration of blocks. 
In our experiments, we treat the task description as an anomalous problem that requires resolution, 
demonstrating the generalization capabilities of the proposed method. 
The experimental results for various block configurations (3, 5, and 7 blocks) are summarized in \Cref{table_llm_blocksworld}. 
Our method exhibits a significantly higher success rate in planning compared to the other methods. 
Moreover, as the number of blocks increases, the complexity of the problem rises; 
however, the success rate of our method decreases at a much slower rate than that of the other methods.

These comparative experiments confirm our hypothesis: by excluding extraneous contextual information and focusing directly on anomalies,
the LLM module can reason more effectively to address the issue.
Additionally, using FLP in the follow-up prompt results in more reliable PDDL task goals.
% On the other hand, in cases where the LLM module fails to generate effective PDDL goals, such as in the {\itshape Object Loss Anomaly} scenario,
% the LLM tends to propose repeated searches of a specific area, potentially leading the robot into a deadlock.
% In the {\itshape Action Blocking Anomaly} scenario, common issues include syntax errors in the constructed PDDL goals, even with the use of few-shot and chain-of-thought prompting methods.
% These problems are more frequent with the GPT-3.5-turbo model, and the quality of the FLP in the {\itshape Action Blocking Anomaly} scenario is often suboptimal.
% However, employing the FLP method alleviates these issues to some extent, depending on the quality of the FLP used.

\begin{table}[htbp]
        \caption{Comparison of different prompting methods in the blocks world experiments.}
        \centering
        \label{table_llm_blocksworld}
        \resizebox{\linewidth}{!}{
                \begin{threeparttable}
                        \begin{tabular}{c|c|c|c|c}% 其中，tabular是表格内容的环境；c表示centering，即文本格式居中；c的个数代表列的个数
                                \toprule %[2pt]设置线宽      
                                \multirow{2}*{Method}  &\multicolumn{3}{C{3.8cm}|}{Success rate}      & \multirow{2}*{Description}                 \\ %换行
                                \cline{2-4}
                                % \midrule %[2pt]  
                                ~ &3 blocks                           & 5 blocks                                   & 7 blocks   & ~   \\
                                \midrule
                                llm-ic  &\textbf{1.0}                           & 0.4                                   & 0.541   & Using GPT-4 as planner with context example \cite{b15.2}   \\
                                \midrule
                                CoPAL  &0.83                            & 0.73                                   & 0.625   & Backprompt after fully executed plan and evaluate goal using GPT-4 \cite{b15.3}   \\
                                \midrule
                                Ours &\textbf{1.0}                            & \textbf{1.0}                                   & \textbf{0.75}   & Using GPT-4 as planner with FLP   \\
                                % \midrule
                                \bottomrule %[2pt]     
                        \end{tabular}
                        \begin{tablenotes}
                                \item[] The bold entities indicate the best results.
                        \end{tablenotes}
                \end{threeparttable}

        }
\end{table}

\subsection{System Implementation}

\subsubsection{Implement in Simulation}

We implemented LLM-PAS in a simple {\itshape Object Loss Anomaly} simulation scenario.
Previous tests using PDDLStream were conducted with a PR2 robot in a Pybullet environment, but since its sensors were not set with corresponding errors, the execution was idealized.
To provide a more realistic assessment, we set up the experimental environment in Gazebo, validating LLM-PAS using a mobile manipulation platform.

The mobile manipulation platform consists of a Husky mobile base, a 7-DoF Franka Emika Panda manipulator, and a standard gripper equipped with a Realsense camera.
The high-level planning tasks are handled by the PDDLStream planner, while the low-level motion planning is managed by MoveIt! and the {\itshape move\_base} packages in ROS.
CSubBT executors are constructed using version 4.0.2 of the BehaviorTree.CPP\footnote{[Online]. Available: https://www.behaviortree.dev/} library.
To integrate the planning-related samplers from PDDLStream into the C++ environment, we used Boost.Python for interfacing between the different compilation platforms.
The actions containing geometric information, obtained from the PDDLStream planner, are automatically transformed into XML files to construct the corresponding CSubBTs.

\begin{figure}[ht]
        \centerline{\includegraphics[width=1.0\columnwidth]{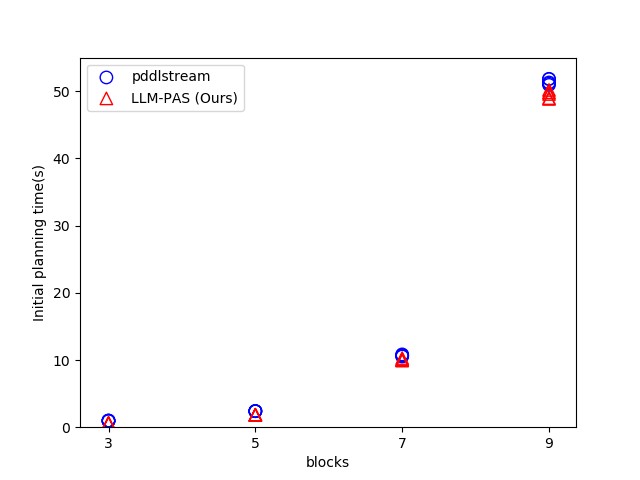}}
        \caption{The initial planning time over 5 trials per problem size.}
        \label{fig_planning_time_exp}
\end{figure}

Before the system experiments, we measured the initial planning time of LLM-PAS and PDDLStream. 
All trials were conducted on a 2.3 GHz Intel Core i7 processor.
As shown in \Cref{fig_planning_time_exp}, LLM-PAS consumes less planning time than PDDLStream, 
as it shifts some constraint-checking tasks from the initial planning phase to execution.
This may not seem to bring about optimization in overall planning time,
but due to the widespread presence of sensor errors during the execution phase, 
it is meaningless to generate some detailed motion plans during the initial planning period, such as the trajectory of manipulation after a base movement.
Therefore, transferring this task to the execution phase helps reduce unnecessary planning time.
and as the complexity of the problem increases or the number of replanning iterations increases, 
LLM-PAS can further reduce the overall planning time.

\begin{figure}[ht]
        \centerline{\includegraphics[width=1.0\columnwidth]{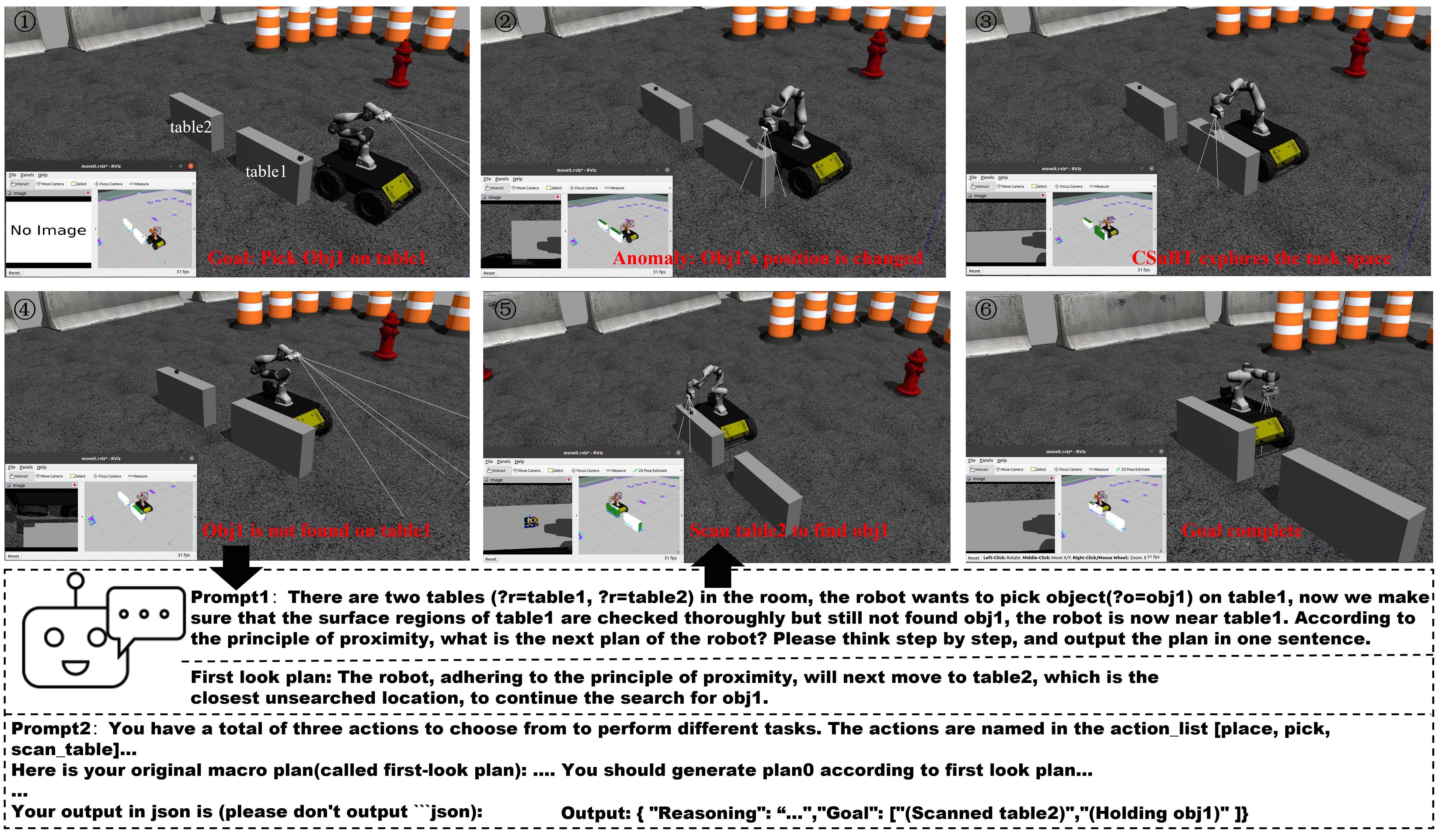}}
        \caption{The systematic experiments of LLM-PAS in simulation.}
        \label{fig_framework_example}
\end{figure}

\Cref{fig_framework_example} illustrates how LLM-PAS operates.
In this scenario, the robot is initially tasked with picking up a cube from table1. However, after the system generates its plan, the cube's location is changed to table2.
The CSubBT, created from the original plan, searches the task space by moving the arm or the base around table1 when the cube is not detected.
If the target remains undetected after sufficient exploration, CSubBT sends a formatted anomaly report to the LLM module.
The LLM module combines the anomaly information with other relevant data to assist the planner in replanning, using the FLP prompting method.
Ultimately, the robot successfully picks up the cube from table2 after the LLM module provides corrected PDDL goals to the planner.

\subsubsection{Implement in Real-world}

\begin{figure}[!h]
        \centerline{\includegraphics[width=\columnwidth]{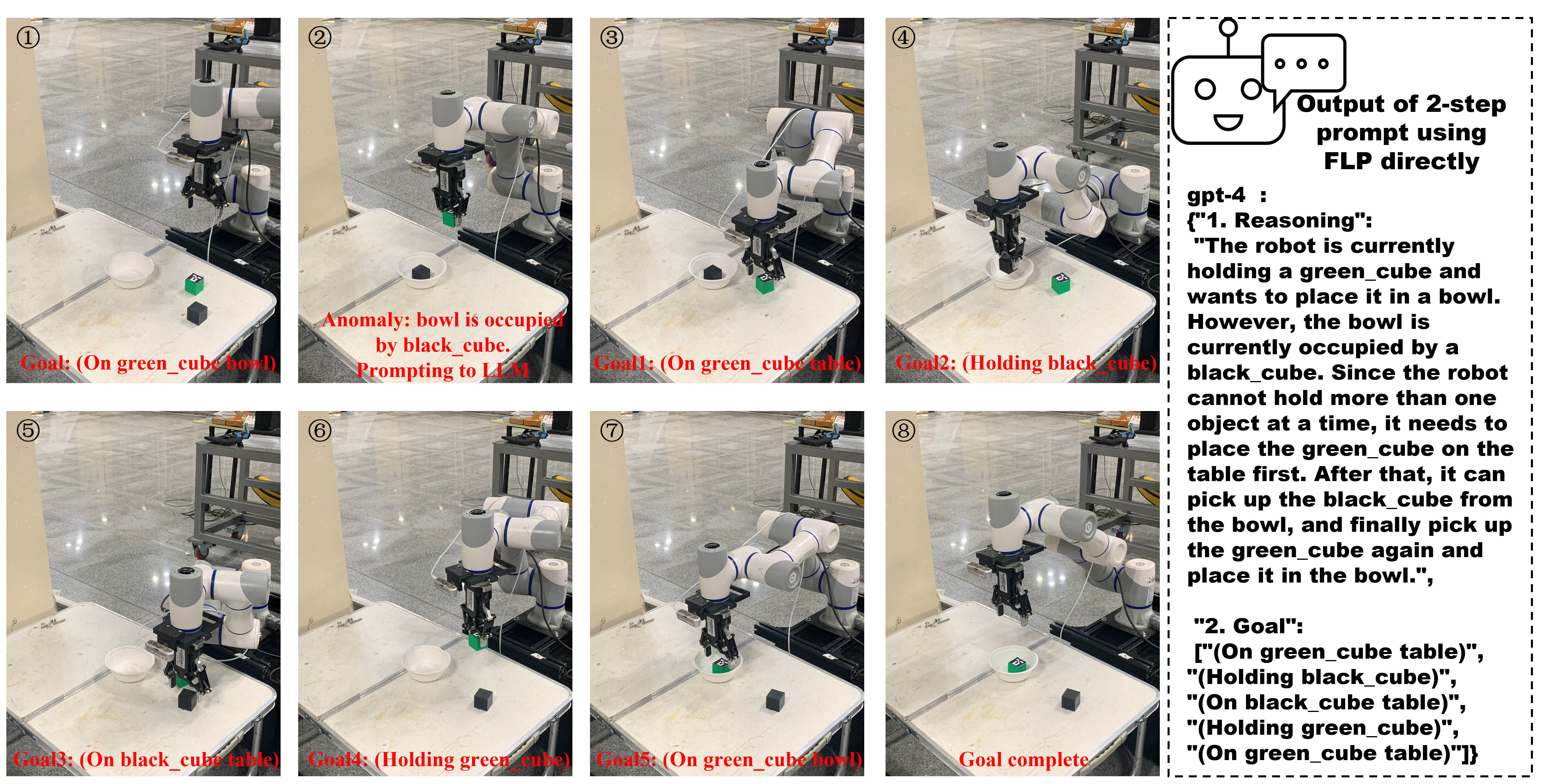}}
        \caption{The systematic experiments of LLM-PAS in real-world.}
        \label{fig_real_world_exp}
\end{figure}

In this experiment, we implemented our system in a simple {\itshape Action Blocking Anomaly} scenario, as shown in \Cref{fig_real_world_exp}.
The setup includes a DOBOT CR3 6-axis robotic arm and a DH-ROBOTICS AG-95 hand gripper.
The robot's initial task was to place a green cube into a bowl.
However, when the robot picked up the green cube and attempted to place it in the bowl, it discovered that the bowl was already occupied by a black cube.
This anomaly information was then processed by the LLM module using the FLP prompting method.
The LLM module reasoned through the situation and generated detailed PDDL task goals to assist the planner in completing the necessary replanning.

The successful execution of these experiments validates the effectiveness and robustness of LLM-PAS.
By leveraging the reasoning capabilities of the LLM module, the system introduces a generalizable replanning mechanism into the traditional TAMP planner.
This enables the task execution system to recover from unexpected anomalies and continue toward task completion, even when the anomaly cannot be resolved by the executor during task execution.

\section{CONCLUSIONS}

In this paper, we introduce LLM-PAS, a closed-loop planning and acting system for TAMP applications, assisted by pre-trained LLMs.
LLM-PAS is designed not only to handle general TAMP tasks but also to enhance the task execution process
by shifting some constraint-checking responsibilities from the planning phase to the actual execution phase.
% This approach enables the system to execute tasks more robustly, without relying on CWA,
% and provides precise feedback to the planner by actively exploring the constraint space when anomalies arise.
Leveraging the LLM's general reasoning capabilities makes LLM-PAS able to address anomalies through replanning,
and we further propose the First Look Prompting (FLP) method to enhance the LLM's stability in generating correct PDDL goals, assisting the planner with the replanning process.
We validate LLM-PAS through both simulation and real-world experiments, demonstrating its effectiveness in handling common anomalies in TAMP tasks.

While our current implementation has shown promise, it is not yet fully optimized, and the cross-platform integration is temporary.
Besides, multimodal LLMs represent one of the most cutting-edge areas in artificial intelligence, offering significant potential for enhancing task planning and acting systems.
As a result, there still remains many work to be done in refining our system and extending its applicability to more diverse scenarios.


%%%%%%%%%%%%%%%%%%%%%%%%%%%%%%%%%%%%%%%%%%%%%%%%%%%%%%%%%%%%%%%%%%%%%%%%%%%%%%%%

%%%%%%%%%%%%%%%%%%%%%%%%%%%%%%%%%%%%%%%%%%%%%%%%%%%%%%%%%%%%%%%%%%%%%%%%%%%%%%%%



% \section*{ACKNOWLEDGMENT}
% We would like to thank OpenAI for providing the GPT models, which serve as the technical foundation for this work.

%%%%%%%%%%%%%%%%%%%%%%%%%%%%%%%%%%%%%%%%%%%%%%%%%%%%%%%%%%%%%%%%%%%%%%%%%%%%%%%%

\end{document}